# Investigating the Impact of COVID-19 on Education by Social Network Mining

***Mohadese Jamalian[a], Hamed Vahdat-Nejad[a*], Hamideh Hajiabadi[b]***

[a] PerLab, Faculty of Electrical and Computer Engineering, University of Birjand, Birjand, Iran

[b] Department of Computer Engineering, Birjand University of Technology, Iran

Mohadesejamalian@birjand.ac.ir, vahdatnejad@birjand.ac.ir, hajiabadi@birjandut.ac.ir

## Abstract

The Covid-19 virus has been one of the most discussed topics on social networks in 2020 and 2021 and has affected the classic educational paradigm, worldwide. In this research, many tweets related to the Covid-19 virus and education are considered and geo-tagged with the help of the GeoNames geographic database, which contains a large number of place names. To detect the feeling of users, sentiment analysis is performed using the RoBERTa language-based model. Finally, we obtain the trends of frequency of total, positive, and negative tweets for countries with a high number of Covid-19 confirmed cases. Investigating the results reveals a correlation between the trends of tweet frequency and the official statistic of confirmed cases for several countries.

## 1. Introduction

Covid-19, as the main phenomenon in 2020 and 2021, has greatly affected the education paradigm. Today, we see many users' comments about the influence of Covid-19 on education on various social networks, especially Twitter. By analyzing the tweets related to the Covid-19 and education, people's feelings about the effects of the disease on education can be extracted. In this regard, via collecting and analyzing related tweets, Australian people's views on homeschooling during Covid-19 have been extracted and their feelings have been classified into six categories, including positive, negative, humor, appreciation of teachers, government/politician opinions, and definitions [1].

In continue, this research aims to investigate the effect of Covid-19 on education from the viewpoint of Twitter users in various countries. To this end, tweets related to the Covid-19 virus are investigated within the early stages of the global outbreak (March to June of 2020). A complete dictionary of words related to education is proposed, using the Oxford[1] Dictionary. Then, using the dictionary, we extract the tweets related to education. To determine the location of the tweets, we create a dictionary of place names using the Geo Names[2] geographic database. In fact, a dictionary-based method is also proposed to determine the location of tweets. Finally, we analyze the sentiment of related tweets using the RoBERTa [2] language-based model and then obtain and analyze the trend of frequency of positive and negative tweets for ten countries[3].

The structure of the remainder of the article is as follows. In the second section, related research is discussed. Section 3 presents the proposed method, and section 4, describes the experiment and outcome. Finally, in the fifth section, concluding remarks are stated.

## 2. Related work

1 https://www.oxfordreference.com/view/10.1093/acref/9780199237043.001.0001/acref-9780199237043

2 https://coronavirus.jhu.edu/map.html

3 USA,India,UK,China,Pakistan,Italy,Australia,Sweden,Japan,Brazil

Today, people use various social networks, especially Twitter, to share their views on various issues and events. By analyzing the published tweets on various topics, implicit and high-level information can be obtained. To this end, tweets should be extracted first and then analyzed according to the research purpose. In this regard, research has been conducted for various purposes, such as extracting users' food preferences [3], extracting tourists' tourism preferences [4, 5], and extracting Syrian refugees' feelings [6].

The Covid-19 virus is the subject of a significant number of tweets published in 2020 and 2021. People's feelings about this virus can be extracted by analyzing tweets related to this virus. Several studies have analyzed tweets related to the Covid-19 virus, which are described below.

The sentiment of the tweets related to the Covid-19 has been analyzed by the TextBlob library in Python [7]. Similarly, a collection of tweets has been collected and analyzed to detect people's emotions about the Covid-19 virus and discover recurring patterns. The emotions are then classified into ten categories, including positive, negative, anger, anticipation, disgust, fear, happiness, sadness, surprise, and trust [8].

In the context of education, 10,421 tweets have been collected to analyze the Australian public's perception of homeschooling during Covid-19 for the three weeks from April 13 to May 3, 2020. Then, the content of the collected tweets has been analyzed, and their sentiments have been classified into six categories, including positive, negative, humor, appreciation of teachers, government/politician opinions, and definitions [1].

In continue, we use the dictionary-based method to extract the tweets related to education in three months and analyze the emotions of tweets separately for various countries.

## 3. Extracting and analyzing Tweets

This study aims to analyze Twitter users' feelings about the impact of the Covid-19 virus on education. To this end, many tweets related to Covid-19 have been considered. Using the dictionary-based method, we extract the tweets related to education and obtain their location. Finally, we analyze the extracted tweets to identify positive and negative sentiments regarding users' emotions.

Since the Covid-19 virus has been prevalent in most countries since late March 2020, the tweets are considered between March 23 and June 23. The keywords "corona", "coronavirus", "covid", "pandemic", "sarscov2" and "covid-19" have been used to identify tweets related to the Covid-19 virus. Next, we use a dictionary-based method to extract tweets related to education. For this purpose, we use the Oxford[4] Dictionary and the help of an expert to create the Education Dictionary. The Education Dictionary contains 134 words related to education. With the help of this dictionary and the implementation of a GATE pipeline, we extract tweets related to education.

Since we intend to analyze the impact of the Covid-19 virus on education for each country separately, the location of each tweet must be determined. We compile a list of place names using the GeoNames[5] geographic database, including states, provinces, and cities in a list of 32 countries[6] selected based on the prevalence of the Covid-19 virus. We use this list in a GATE pipeline [9] as a Gazetteer list to match the tweets with the list, thus specifying a location tag for each tweet.

We propose an emotion classification model based on the RoBERTa language-based model [2]. Since RoBERTa is a new model with impressive results, it is used for sentiment analysis. In order to train the sentiment classification model, three labeled datasets including Stanford Sentiment treebank (67,300

[4] https://www.oxfordreference.com/view/10.1093/acref/9780199237043.001.0001/acref-9780199237043

[5] http://download.geonames.org/export/dump

[6] Australia, Belgium, Brazil, Canada, Chile, China, Ecuador, France, Germany, India, Iran, Ireland, Italy, Japan, Mexico, Netherlands, New Zealand, Pakistan, Peru, Portugal, Qatar, Russia, Saudi Arabia, Singapore, South Korea, Spain, Sweden, Switzerland, Turkey, UAE, UK, USA

samples) [10], SemEval 2015 Task 10 (6800 samples) [10] and SemEval 2015 Task 11 (3500 samples) [11] were used. The RoBERTa model receives a tweet text (string) as input and then provides a set of token representations $h_t(t = 1, \dots, T)$ as output. In the following, we consider three Separate classifications for each of the three datasets. Each classifier receives a vector representation of the tweet as input. This vector representation, called H, is obtained by the following formula:

$$H = \sum_t e_t * h_t$$

Where

$$e_t = \frac{e^{\alpha_t}}{\sum_j \alpha_j}, \qquad \alpha_t = w_{att} h_t$$

Finally, in the test phase, with the help of the majority vote from the three categories, the final tag of the tweet is determined. The labeling of the RoBERTa model is such that it assigns a zero score to tweets with negative content and a score of 1 to tweets with positive content. In this way, the sentiments of tweets are divided into two categories: positive and negative.

Once we have obtained the sentiment score of the tweets using the RoBERTa language-based model, we can calculate the frequency of positive and negative tweets per week for each country. In this way, we can acquire temporal information about the opinions and thoughts of people in different countries about the Covid-19 virus and the field of education.

## 4. Experiment

We have used GATE software [12] to extract tweets related to education from a dataset involving millions of tweets related to the Covid-19 virus, which were posted from March 23 to June 23, 2020. GATE software can process any language and develop the components for processing human language [13]. We use the location dictionary in a GATE pipeline as a Gazetteer list to match the tweets with the list, thus specifying a location tag for each tweet. The tagging is done so that for each occurrence of the name of a city or state, the name of that country is attributed to the tweet.

After analyzing the sentiments of the tweets, the frequency diagrams of all tweets, positive tweets, and negative tweets for ten countries with a sufficient number of related tweets have been drawn separately. Also, we acquired the official statistics of the Covid-19 confirmed cases[7] for the investigated countries and added them to the charts. Figure 1 includes the frequency of total tweets, positive tweets, and negative tweets per week, as well as the official Covid-19 confirmed cases for each of the selected countries. Each diagram includes two vertical axes; the left axis shows the formal daily confirmed cases of the Covid-19 virus, and the right axis indicates the frequency of tweets.

[7] https://coronavirus.jhu.edu/map.html

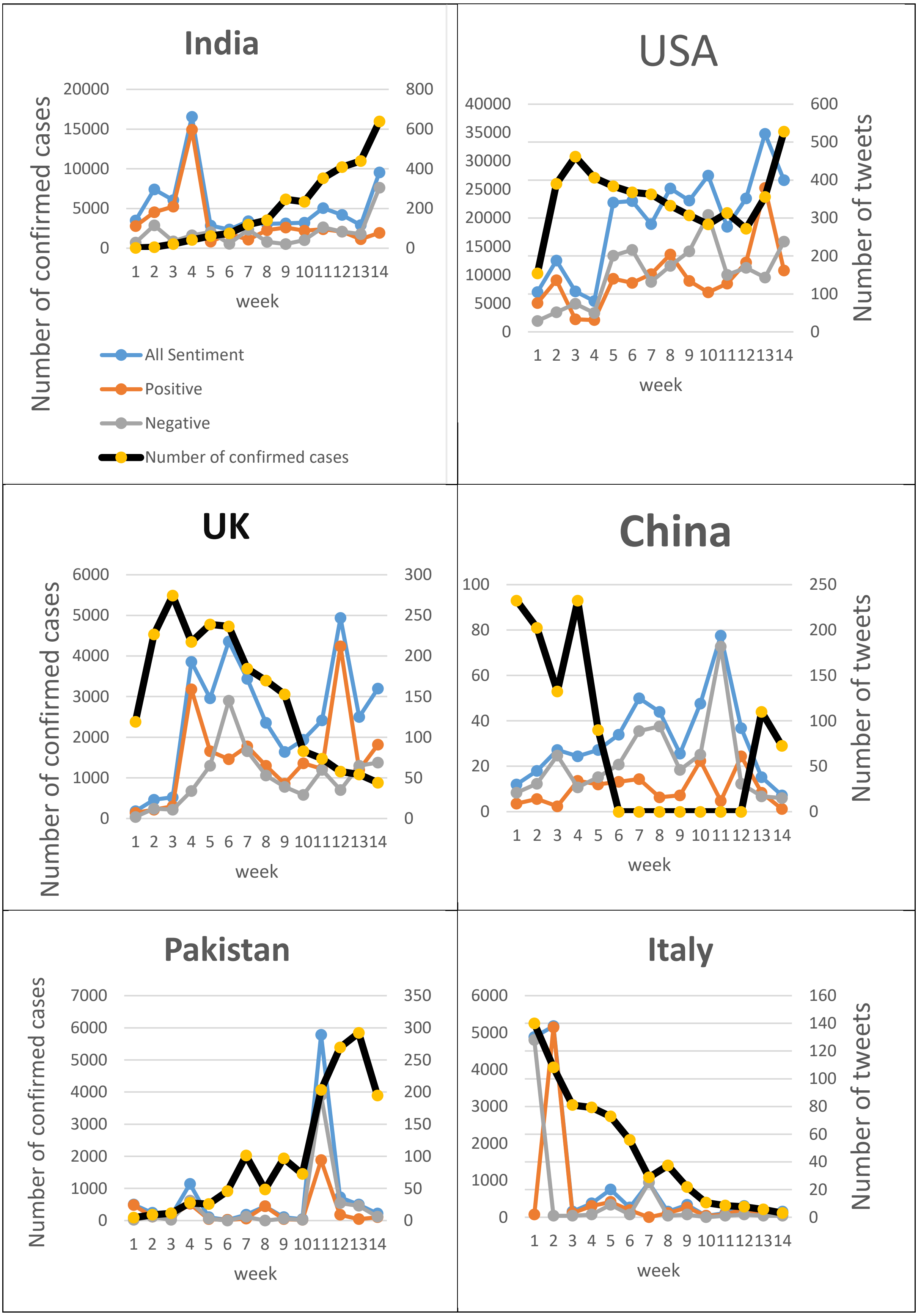
India
Number of confirmed cases
20000
15000
10000
5000
0
800
600
400
200
0
1 2 3 4 5 6 7 8 9 1011121314
week
All Sentiment
Positive
Negative
Number of confirmed cases
USA
40000
35000
30000
25000
20000
15000
10000
5000
0
600
500
400
300
200
100
0
Number of tweets
1 2 3 4 5 6 7 8 9 1011121314
week
UK
Number of confirmed cases
6000
5000
4000
3000
2000
1000
0
300
250
200
150
100
50
0
1 2 3 4 5 6 7 8 9 1011121314
week
China
100
80
60
40
20
0
250
200
150
100
50
0
Number of tweets
1 2 3 4 5 6 7 8 9 10 11 12 13 14
week
Pakistan
Number of confirmed cases
7000
6000
5000
4000
3000
2000
1000
0
350
300
250
200
150
100
50
0
1 2 3 4 5 6 7 8 9 1011121314
week
Italy
6000
5000
4000
3000
2000
1000
0
160
140
120
100
80
60
40
20
0
Number of tweets
1 2 3 4 5 6 7 8 9 1011121314
week

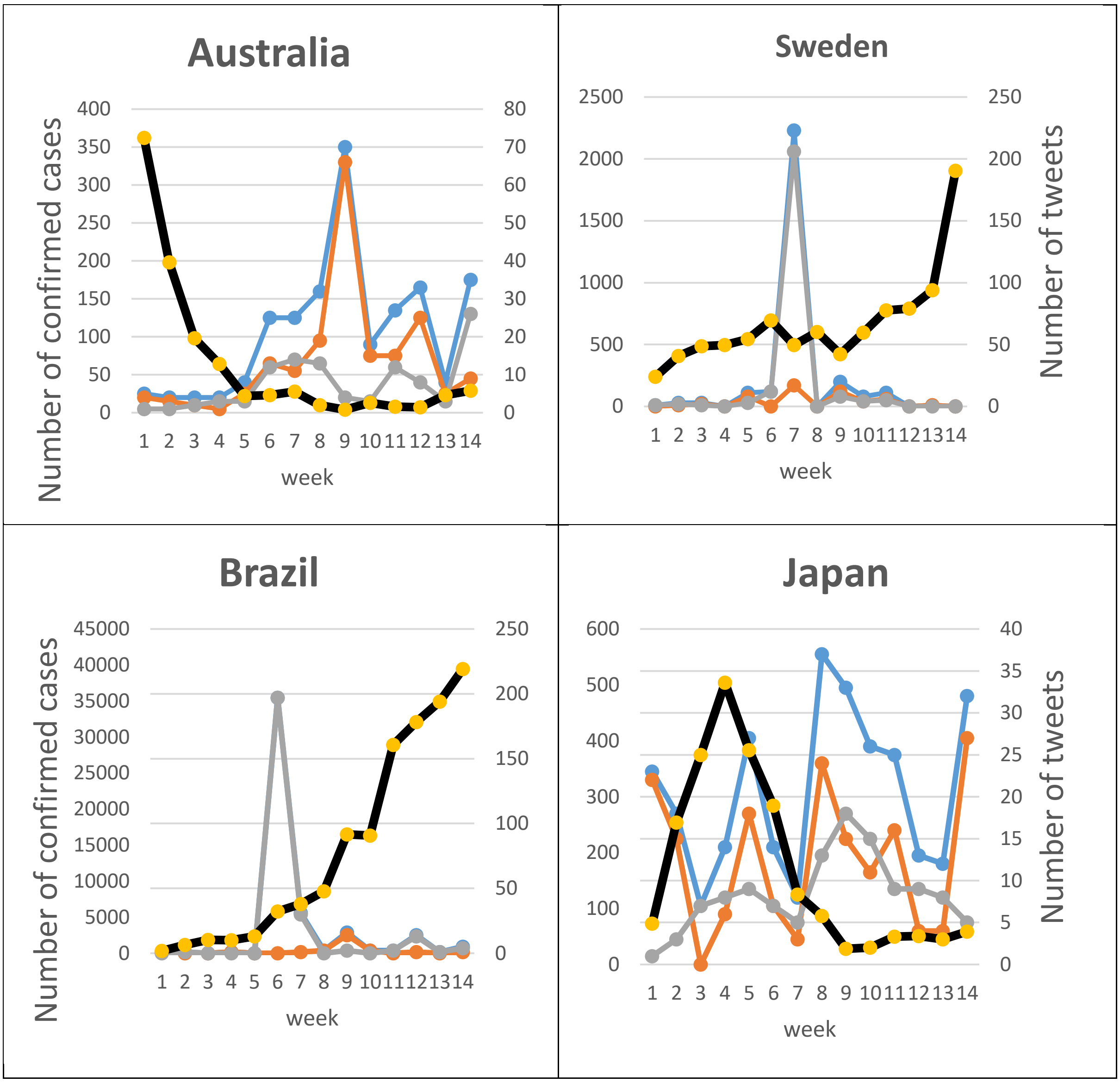

Figure 1. Frequency of total, positive, and negative tweets, as well as official daily confirmed cases of Covid-19

Comparing the frequency trend of all the tweets with the official statistics curve of the Covid-19 reveals a correlation between these two curves for some countries such as the United States and Pakistan. Besides, it can be seen that in India and Italy, the number of negative tweets increases at the peak of the official statistic of the Covid-19 virus. Also, the number of positive tweets is high when the official statistic becomes low.

In some investigated countries such as Australia, India, England, and Italy, positive tweets are roughly more than negative tweets. In contrast, in other countries such as China, Pakistan, and the United States, the average number of negative tweets is higher than positive tweets.

## 5. Conclusion

In this research, the attitude of Twitter users regarding the impact of the Covid-19 virus on education has been investigated. We have analyzed related tweets in three months of the initial stages of the global outbreak. The location of the tweets is then determined by creating a location dictionary using the GeoName geographic database and running a GATE pipeline. Then, the sentiments of the tweets were analyzed using the RoBERTa-based model and classified into two categories of positive and negative emotions. Finally, considering the frequency of total, positive and negative tweets for each selected country, we identify a correlation between the frequency of tweets with the official confirmed cases statistics in several countries. In the future, a more specific analysis will be conducted to detect the influence of Covid-19 on any aspect of education, such as remote education, research, etc.